\pdfoutput=1
\documentclass[11pt]{article}
\usepackage[preprint]{acl}

\usepackage{times}
\usepackage{latexsym}
\usepackage{graphicx} 
\usepackage[T1]{fontenc}
\usepackage[utf8]{inputenc}
\usepackage{microtype}
\usepackage{hyperref}
\usepackage{amssymb}
\usepackage{tcolorbox}

\usepackage{tabularx}
\usepackage{booktabs}
\usepackage{graphicx}
\usepackage{listings}
\usepackage{multirow}
\usepackage{enumitem}
\usepackage{rotating}
\usepackage{array}
\usepackage{color}
\usepackage{adjustbox}

\title{Generating Unseen Code Tests In Infinitum}

\author{Marcel Zalmanovici \and Orna Raz \and Eitan Farchi \and Iftach Freund\\ 
         IBM Research AI \\ 
         \small{\texttt{\{marcel, ornar, farchi\}@il.ibm.com}
         \texttt{\{iftach.freund\}@ibm.com}}
         }

\begin{document}
\maketitle

\begin{abstract}
Large Language Models (LLMs) are used for many tasks, including those related to coding. An important aspect of being able to utilize LLMs is the ability to assess their fitness for specific usages. The common practice is to evaluate LLMs against a set of benchmarks.
While benchmarks provide a sound foundation for evaluation and comparison of alternatives, they suffer from the well known weakness of leaking into the training data 
\cite{Xu2024Benchmarking}. We present a method for creating benchmark variations that generalize across coding tasks and programming languages, and may also be applied to in-house code bases. Our approach enables ongoing generation of test-data thus mitigating the leaking into the training data issue. We implement one benchmark, called \textit{auto-regression}, for the task of text-to-code generation in Python. Auto-regression is specifically created to aid in debugging and in tracking model generation changes as part of the LLM regression testing process.


\end{abstract}

\section{Introduction}
Many benchmarks exist for evaluating LLM generation that is related to software development and coding tasks \cite{zhang2023unifying}. Two well acknowledged challenges related to utilizing benchmarks for evaluation are (1) data leakage, where the training data of the LLM under evaluation eventually includes benchmarks' data, and (2) debugging or the ability to understand what the underlying issues highlighted by benchmark failures are so that they can be mitigated. 

We introduce an approach for generating benchmarks for code related tasks that relies on a commonly used intermediate code representation in the form of an Abstract Syntax Tree. ASTs are often used in program analysis. In the context of LLMs they are sometimes used to create metrics that may better fit code-related tasks, including robustness related assessment \cite{astxplainer2024evaluating,astdatasetgen2024robust}. We are unaware of work that utilizes ASTs to automate the generation of benchmarks and ease the debugging process. Our approach makes it feasible to create new benchmarks in order to ensure that the evaluation can generalize as we can be confident that the LLM has not trained on this newly generated benchmark data. Our approach supports the creation of a debugging dictionary that includes programming language constructs that have been identified as challenging for the LLM to correctly generate. Once manually populated, this dictionary can be used to easily debug new LLM output. 

\begin{figure}
    \centering
    \includegraphics[width=1\linewidth]{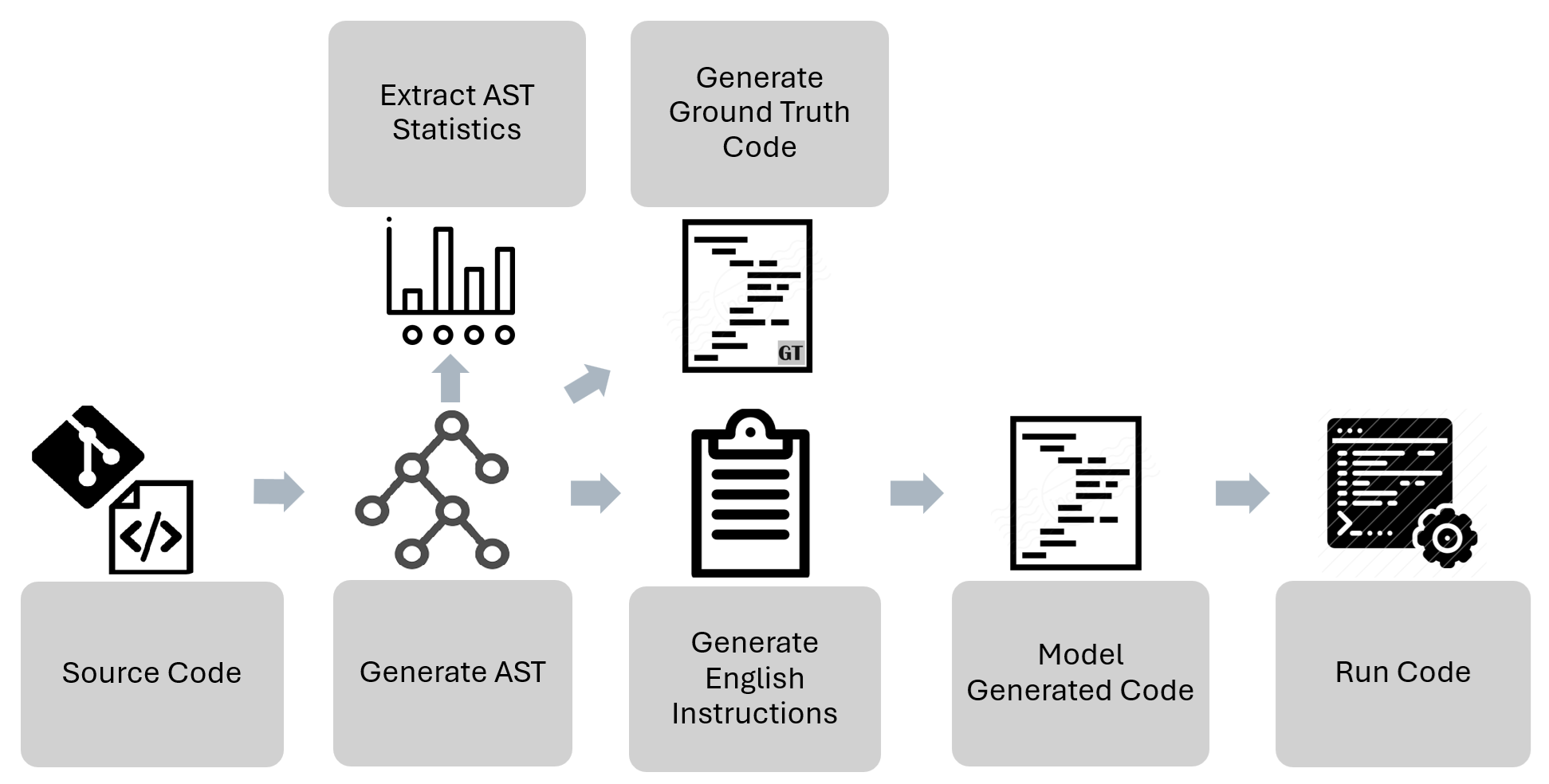}
    \caption{The process of generating custom benchmarks and running them}
    \label{fig:bench_process}
\end{figure}

We introduce our AST-based approach for generating code related benchmarks. 
Figure \ref{fig:bench_process} depicts the process of generating a benchmark and testing it. The process begins by \emph{selecting source code} to test the LLM with. The code should have unit-tests, allowing to verify model output more accurately. It would also preferably include code that was unseen during training. The chosen source code may be in a different programming language than the task destination code.
The next step is to \textit{generate the AST}. Our suggestion is to use a tool like tree-sitter \cite{tree-sitter}. 

We demonstrate the approach depicted in Figure \ref{fig:bench_process} by generating a new benchmark, the \emph{auto-regression benchmark}. The starting point for auto-regression is the NAPS dataset \cite{naps} and the AST representation that it provides (termed universal AST or uAST). 

Auto-regression is a text to code benchmark where each output set of instructions, termed \textit{problem}, includes test cases as input-output tuples. 
The text is generated deterministically from the NAPS AST to ensure its correctness. Auto-regression can be viewed as following the paradigm of IFEval \cite{zhou2023instructionfollowing} and extending it with a low-level textual instructions benchmark for code tasks. It is intended as a basis for regression testing and eliminates the need to debug algorithmic related issues as it provides the correct algorithm to implement. We demonstrate the notion of a debugging dictionary for auto-regression and how we utilize it to easily debug LLM generation issues. Having a single dictionary helps in creating regression tests as it makes it possible to track changes in LLM versions. We see cases where newer LLM versions improve in some categories yet introduce other categories of problems. 


To summarize, this paper's contributions are:
\begin{enumerate}
    \itemsep0em 
    \item  A method for creating benchmarks in a way that generalizes across tasks and programming languages; the benchmark can be created from any existing code, including in-house proprietary code. It is recommended that the code has unit tests.   
    
    \item A low-level text-instructions to code benchmark, the auto-regression benchmark. 
    \footnote{To be released along with the publication}
    We tested Python generation from the benchmark.
    
    \item A method for creating a debugging dictionary that becomes part of the auto-regression benchmark; this dictionary enables regression testing of deterioration or improvement in implementing the dictionary constructs. 
\end{enumerate}

\section{AST-based benchmark generation}

\textit{Generating English instructions} from an AST can be done by walking the tree nodes and outputting a description from the leaves toward the root. An example of the generated list of instructions is in Figure \ref{ex:instructions}. To keep the generation simple we limited ourselves to a single pass over the tree. This means, for example, that we add redundant parenthesis in mathematical expressions to make sure they are not ambiguous. We also leave calls like \(f(g(h()))\) untouched even though, for a human, expanding the calls would be clearer.  For example, \textit{concat\_string(a, concat\_string(b, c))} would be clearer to a human if expanded to \textit{concatenate string a with b and c}.
It is possible to adjust the detail level and expressiveness of the English generated output depending on how well the model performs and what we are trying to test.

To get the \textit{model generated code} we pass the model a set of generic instructions and the text generated in the previous step. The generic instructions are lists of "do" and "don't" clarifications that help improve performance. For example, "Do not ask for user input!" was very important mainly to llama-family models as they tend to add unrequested code asking for user input in the \textit{global scope}, i.e. outside any Python method. This code is executed upon import and, as a result, automated tests would get stuck at this stage. Another example is: "Replace all array\_* methods with Python list operations" which models (except GPT) often left unimplemented even though another instruction explicitly requests the implementation to cover all aspects of the provided pseudo-code. 

Finally, the \textit{running the code} stage uses unit-test assertions to score the performance of the model. We run the models using a \textit{greedy} setting, when possible, to get the best solution the model can come up with. We grade the model performance using a school-like grading method where each test is a sub-problem that earns equal points. For example, if a problem has 8 tests and the model passes 6 it gets a grade of 0.75. We prefer this over the commonly used \textit{pass@k} \cite{chen2021evaluating} because (1) we want a best-effort attempt, not the probability to get at least one correct solution from $k$ attempts, and (2) to save inference time (and cost) of running the model $k$ times.

From the AST we also \textit{generate ground-truth run-able code} in the target language. This code can be used to generate prompts for the fill-in-the-middle (FIM) task, for running model metrics which compare code text \cite{Papineni2002BleuAM, Lin2004ROUGEAP} or build their own code representation \cite{ren2020codebleu, zhou-etal-2023-codebertscore}, and for debugging the model code since due to the low-level instructions we expect little variation in outputs.

Lastly, we also \textit{generate AST statistics}. We use this to identify programming language constructs where the model under evaluation is more likely to err. We also count how many times each construct appears in each problem. The statistic includes basic constructs like if/if-else conditions and loops with and without continue or break keywords. It also includes nesting levels for loops, whether there are operations on data structures such as list/dictionary/set. Finally, motivated by manual analysis of model errors we added statistics about things like ASCII operation.

\subsection{Data}


We used the NAPS dataset \cite{naps} which is generated from solutions for competition problems posted on \hyperlink{www.codeforces.com}{CodeForces}. This dataset was appealing because it represented real-world problems for which there is a high-level description; it has multiple examples of working code; there exist test cases and each problem has tens to hundreds of low-level instructions created by humans.
Unfortunately, the human low-level instructions proved to contain too many errors, therefore we decided to use the AST it provided and write an AST to English translator.
The advantage of translation using code is the consistency in description of the same operation. The downside is a lack of variation which might catch more model issues.

We created our auto-regression datasets in two sizes: a \textit{tiny} one consisting of 135 problems and a \textit{small} one with 460 problems. The available data allows generation of a dataset with a few thousand problems. However, this doesn't provide much additional benefit. On the small dataset we saw that models had a higher success rate than on the tiny one. We did not see new categories of errors, thus concluded that most instructions are easy enough to follow, therefore there is no need to add more tests of the same construct. 

Example \ref{ex:instructions} shows the instructions generated from the AST of a simple programming problem. More examples can be found in Appendix \ref{appendix:code_examples} where we show how common errors look like.

{
    \renewcommand{\tablename}{Figure}
    \begin{table}[ht]
        \centering
            \lstset{
                basicstyle=\ttfamily\small,
                breaklines=true,
                frame=single
            }
            \begin{tiny}
            \begin{lstlisting}
Define a function called __main__ getting as parameters var0 as integer, var1 as integer and returns string.
Declare var2 as integer, var3 as integer, var4 as integer.
Assign (var0 divided by func0(var0, var1)) multiplied by var1 to var2.
Assign var2 divided by var0 to var3.
Assign var2 divided by var1 to var4.
If var0 is greater than var1 then assign var3 plus 1 to var3.
Otherwise assign var4 plus 1 to var4.
If var3 is greater than var4 then return "Dasha"
Otherwise if var3 is less than var4 then return "Masha"
Otherwise return "Equal"
            \end{lstlisting}
            \end{tiny}
        \caption{Example instructions generated from an AST.}
        \label{ex:instructions}
    \end{table}
}


\subsection{Metrics}

Since we have tests cases, we decided to calculate partial and whole test pass/fail scores.
The whole test pass/fail is the same as the popular \textit{pass@k} where $k=1$.
The partial score treats the tests/asserts as test questions; getting all questions right gives a full mark, getting $m$ out of $n$ correct gives a score of $m/n$.

Not all problems in the dataset have the same number of tests. In the \textit{tiny} dataset most problems have 10 tests, but several have less, usually 7 or 5.
In the \textit{small} dataset there are some problems that have 100 or more tests.

\subsection{Debugging dictionary}
Auto-regression is created such that the prompts it generates are an exact description of the desired implementation. This enables to create a dictionary of low level coding constructs that assists in debugging generated code that fails to correctly pass the given tests. This dictionary can be repeatedly used over the benchmark execution results as part of a test regression analysis. 

\section{Benchmark results}

We ran the dataset on GPT, LLama, Mixtral, Granite and Deepseek. For most we also tried more than one version of the model for comparison. The models size varies between 8b and 70b.

Table \ref{tab:models} summarizes the results of running the \textit{tiny} version of the auto-regression dataset on several models.
The table is sorted by \textit{W (Whole)} test results. The clear winner is \textit{gpt-4o} which failed on only 9 programs, followed by \textit{llama-3-70b} and \textit{deepseek}. Since all models did sometimes pass part of the tests (in some cases 7-9 of 10) the \textit{P (Partial)} score is higher for all models.

It is interesting to note that \textit{gpt-3.5-turbo} had many issues with compilation -- all due to unmatched parenthesis, issue completely solved in \textit{gpt-4o}.
Also, the older version, 1106, has less programs that fail to compile -- a regression. The overall score of the newest version, 0125, is slightly better (partial score), meaning the model performed better when the generated program compiled.

The infinite loop issue was mainly caused by "forgetting" to update the loop variable either at the end of the loop or at a continue keyword. 

\begin{table}[htbp]
\centering
\small
\setlength{\tabcolsep}{3pt}
\begin{tabular}{>{\bfseries}p{3.95cm}|>{\centering\arraybackslash}p{0.6cm}>{\centering\arraybackslash}p{0.6cm}>{\centering\arraybackslash}p{0.8cm}>{\centering\arraybackslash}p{0.5cm}}
\toprule
\textbf{Model} & \textbf{W} & \textbf{P} & \textbf{Static Err} & \textbf{Inf Err} \\
\midrule
gpt\_4o                        & 0.93 & 0.95 & 0 & 0 \\
llama\_3\_70b\_instruct        & 0.83 & 0.87 &0 & 0 \\
deepseek\_coder\_33b\_instruct & 0.70 & 0.75 &0 & 5 \\
gpt\_3.5\_turbo\_0125          & 0.67 & 0.73 &11 & 5 \\
gpt\_3.5\_turbo\_1106          & 0.67 & 0.72 &6 & 5 \\
granite\_34b\_code\_instruct   & 0.54 & 0.62 & 2 & 7 \\
granite\_8b\_code\_instruct    & 0.50 & 0.62 & 1 & 7 \\
codellama\_34b\_instruct       & 0.45 & 0.51 & 10 & 9 \\
mixtral\_8x7b\_instruct\_v01   & 0.43 & 0.52 & 3 & 5 \\
llama\_3\_8b\_instruct         & 0.41 & 0.49 & 6 & 11 \\
\bottomrule
\end{tabular}
\caption{Comparison of different models on the \textit{tiny} version (135 programs) of the auto-regression dataset. The \textit{W (Whole)} column is the percent of programs which passed all tests; the \textit{P (Partial)} the percent of tests passed (not all programs have the same number of tests); \textit{Static Err} are programs that didn't parse; \textit{Inf Err} are programs stuck in infinite loops}
\label{tab:models}
\end{table}

Table \ref{tab:error_analysis} summarizes the errors in most of the models we analyzed and categorizes them. Note that:
\begin{itemize}
    \item Each program may have multiple issues. Therefore, summing the row may be different than the overall error rate.
    \item Some issues could have been assigned to more than one category, e.g. ignoring string split could be both ignored (chosen label) or split. 
    \item It is possible that not all issues in a program were logged, especially for programs that didn't compile or when one or more instructions were missing.
    \item Some programs passed several or even most tests even though they contained errors.
\end{itemize}

There are two generic errors, \textit{ignored} -- the model didn't implement some instruction or implemented only part of it, and \textit{wrong} -- the model implemented something other than what was asked of it, a hallucination.
There are several errors which were repeated often enough and across models that they received their own category. More details on the meaning of the error labels are in Appendix \ref{appendix:error_labels}.
To test the model generated code we ran it on the tests. We manually analyzed the failures and identified the problem. To facilitate the analysis we used the statistics to focus on potential issues. For example, after discovering that the models had a recurring issue with updating the loop variable when the \textit{continue} keyword should be used, if a test failed and the stats mentioned \textit{continue} as a construct we first checked whether that was the issue. 

The results show that the performance of GPT increased significantly between the 3.5 turbo and 4o versions. The 3.5 had a major issue with closing parenthesis, mainly when there were a lot of them in the instructions.
Anecdotally, a previous, deprecated, version of 3.5 didn't suffer from that issue and we measured a performance around 0.8. 

For Granite the small 8b model couldn't implement string split correctly, an issue that was mostly solved by the newer and larger 34b model. The latter also improved in handling of loops and ASCII. Interestingly, it "replaced" the wrong implementation with ignoring part/whole instructions. We need to verify whether the errors are on the same constructs/instructions in both versions.

\begin{table}[htbp]
\centering
\small
\setlength{\tabcolsep}{3pt}
\begin{tabular}{>{\bfseries}p{1.6cm}p{0.5cm}p{0.5cm}|p{0.5cm}p{0.5cm}|p{0.5cm}|p{0.5cm}p{0.5cm}}
\toprule
 & \multicolumn{2}{c}{\tiny Granite} & \multicolumn{2}{c}{\tiny GPT} & \multicolumn{1}{c}{\tiny Deepseek} & \multicolumn{2}{c}{\tiny Llama} \\ 
 \cmidrule(l){2-8}
 & \tiny 8b & \tiny 34b & \tiny 3.5 & \tiny 4o &  & \tiny v3 70b & \tiny code 34b  \\ 
\hline
ignored     & 14 & 22 & 9  & 0  & 8  & 2 & 12 \\
wrong       & 39 & 11 & 15 & 8  & 22 & 5 & 23 \\
loop        & 13 & 6  & 5  & 0  & 18 & 1 & 10 \\
ASCII       & 13 & 9  & 3  & 0  & 3  & 3 & 15 \\
unbalanced  & 0  & 1  & 11 & 0  & 0  & 0 & 3  \\
division    & 2  & 8  & 3  & 1  & 3  & 2 & 4  \\
indent      & 8  & 8  & 1  & 0  & 1  & 1 & 5  \\
split       & 15 & 5  & 0  & 0  & 0  & 7 & 4  \\
global      & 2  & 2  & 2  & 0  & 3  & 2 & 2  \\
other       & 0  & 2  & 0  & 0  & 0  & 0 & 9  \\
\bottomrule
\end{tabular}%
\caption{The distribution of the errors on the dataset. The first two rows represent instructions that were partially or completely ignored (not implemented) or "misunderstood" (hallucinations); some errors could be placed in either. All other rows -- our dictionary of issues, represent specific errors repeated often, usually across models.}
\label{tab:error_analysis}
\end{table}

\section{Conclusions}
We introduced an AST-based methodology for automatically generating benchmarks for LLM code related tasks. We utilized this approach to generate the auto-regression benchmark, a low level instructions text to Python code benchmark. Relying on ASTs also allowed us to include a dictionary of Python constructs to ease the debugging task. Our results provide an anecdotal indication for the usefulness of our approach and benchmark in overcoming 
two well acknowledged challenges related to benchmarks -- data leakage and debugging. 

\newpage
\section{Limitations}
We believe our methodology could apply to multiple tasks and programming languages (PLs). However, in the paper we present only a single task and programming language. We need to implement more benchmarks from a variety of code repos, for a variety of tasks and PLs. 

The ability to create a debugging dictionary may depend on the task and LLMs tested. It might also depend on the level of details provided in the prompts. We need to test our approach when implementing benchmarks with varying levels of detail. 

In the work presented here we relied solely on dynamic code execution metrics. Relying on such metrics may be too strong a requirement and it would be good to add static metrics. 
\bibliography{main}

\onecolumn
\onecolumn 
\appendix

\section{Appendix A: Code Examples}
\label{appendix:code_examples}

Below is an example of the full list of the low-level instructions the models were asked to implement. The instructions always start with the function name and parameters, followed by the list of local variables (remnant of C/C++ in the AST) and then the instructions themselves.

The model was asked to take a C-style while loop, which is uncommon in Python where for-each is prevalent, and implement it in Python. The loop updates $var7$; it also has a condition which $continues$ to the next iteration.

\begin{tiny}
\begin{lstlisting}[breaklines=true,escapechar=!]
Define a function called __main__ getting as parameters var0 as integer, var1 as list and returns integer.
Declare var2 as integer, var3 as integer, var4 as integer, var5 as list of strings, var6 as integer, var7 as integer, var8 as integer, var9 as integer, var10 as integer, var11 as integer, var12 as integer.
Assign len(var1) to var2.
Assign var0 to var3.
Assign 0 to var4.
While var4 is less than var2 do:
	1. Assign var4 plus 1 to var4.
Assign a new list with dimensions of sizes var2, var3 to var5.
Assign 0 to var6.
Assign var2 minus 1 to var7.
!\colorbox{yellow}{While var7 is greater than or equal to 0 do the following and decrement var7:}!
	1. Assign var3 minus 1 to var8.
	2. While var8 is greater than or equal to 0 do the following and decrement var8:
		1. Assign (1 if array_index(array_index(var1, var7), var8) is equal to 87 else 0 minus 1) to var9.
		2. Assign var9 minus array_index(array_index(var5, var7), var8) to var10.
	!\colorbox{yellow}{3. If var10 is equal to 0 then continue to the next iteration.}!
	4. Assign var6 plus 1 to var6.
	5. Assign 0 to var11.
		6. While var11 is less than or equal to var7 do the following and increment var11:
			1. Assign 0 to var12.
	2. While var12 is less than or equal to var8 do the following and increment var12:
		1. Assign array_index(array_index(var5, var11), var12) plus var10 to array_index(array_index(var5, var11), var12).
Return var6   
\end{lstlisting}
\end{tiny}

In this case, the model correctly implemented the update for the loop itself, but forgot to do the same before the continue keyword. 
This type of error happened with most models. It is interesting to note that although there are 7 programs in the data-set with the continue keyword, the models got this wrong 1-3 times and correct the rest of the times.

\begin{tiny}
\begin{lstlisting}[breaklines=true,escapechar=!]
def __main__(var0, var1):
    var2 = len(var1)
    var3 = var0
    var4 = 0
    var5 = [[0 for i in range(var3)] for j in range(var2)]
    var6 = 0
    var7 = var2 - 1
    while var7 >= 0:
        var8 = var3 - 1
        while var8 >= 0:
            var9 = (1 if var1[var7][var8] == 87 else 0) - 1
            var10 = var9 - var5[var7][var8]
            !\colorbox{yellow}{if var10 == 0:}!
                !\colorbox{yellow}{continue}!
            var6 += 1
            var11 = 0
        while var11 <= var7:
                var12 = 0
                while var12 <= var8:
                    var5[var11][var12] += var10
                    var12 += 1
                var11 += 1
            var8 -= 1
        !\colorbox{yellow}{var7 -= 1}!
    return var6   
\end{lstlisting}
\end{tiny}

To save space, for the rest of the examples we will show partial snippets of both instructions and resulting code.

Another common error source is ASCII handling. Here the model is asked to implement a function which operates on the integer representation of ASCII chars. It is asked to create a utility method that add/subtracts from a char returning the new char and then use it. 

\begin{tiny}
\begin{lstlisting}[breaklines=true,escapechar=!]
Define a function called func0 getting as parameters !\colorbox{yellow}{var0 as character and returns character}!.
Return (var0 plus -48 if (var0 is greater than or equal to 48) and (var0 is less than or equal to 57) ...

Define a function called __main__ getting as parameters !\colorbox{yellow}{var1 as string}! and returns string.
...
While var6 is less than var2 do the following and increment var6:
	 1. If array_index(var1, var6) is not equal to 61 then assign var3 plus (func0(array_index(var1, var6)) multiplied by var5) to var3.
	 2. Assign var5 minus 1 to var5.
\end{lstlisting}
\end{tiny}

The model correctly implemented the utility method calling $ord$ and $chr$ where needed, but failed to do so in the main function.

\begin{tiny}
\begin{lstlisting}[breaklines=true,escapechar=$]
def func0(var0):
    if $\colorbox{yellow}{48 <= ord(var0) <= 57}$:
        return $\colorbox{yellow}{chr(ord(var0) - 48)}$
...

def __main__(var1):
...
    while var6 < var2:
        if $\colorbox{yellow}{var1[var6]}$! != 61:
            var3 += $\colorbox{yellow}{func0(var1[var6])}$ * var5
        var5 -= 1
        var6 += 1
\end{lstlisting}
\end{tiny}

And here is an example where the model misinterpreted a simple assignment, one out of more then 10 in the same program.

\begin{tiny}
\begin{lstlisting}[breaklines=true,escapechar=$]
Define a function ...
...
Assign len(var0) to var2.
...
Assign 0 to var8.
While var8 is less than var2 do the following and increment var8:
	 $\colorbox{yellow}{1. Assign var9 to var7.}$
  ...
\end{lstlisting}
\end{tiny}

The assignment was inverse than the requirement. It is clear that the model understand the sentence \textit{assign x to y} as \textit{y = x} since it does so correctly most of the time in this and other programs, however, there are a few cases where this goes wrong.

\begin{tiny}
\begin{lstlisting}[breaklines=true,escapechar=$]
def __main__(var0, var1):
...
    var2 = len(var0)
...    
    var8 = 0
    while var8 < var2:
        $\colorbox{yellow}{var9 = var7}$
...        
\end{lstlisting}
\end{tiny}

A different kind of error, ignoring part of or the whole instruction, is also common. Below is an example which might be explained by the model training data.
The model skips implementing the update of \textit{var6}.

\begin{tiny}
\begin{lstlisting}[breaklines=true,escapechar=$]
Define a function called func0 ...
...
Assign var2 to var5.
Assign a new list of integers to var6.
$\colorbox{yellow}{array\_push\\(var6, -1\\)}$
Assign 0 to var7.
\end{lstlisting}
\end{tiny}

If the instructions changing the value of the same variable are separated by a few other instructions, e.g. in this case the new $var6$ initialization pushed up a couple of instructions, then the model correctly implements the whole sequence.
This might be caused by the fact that typically programmers do not define a variable and immediately on the next row update it.

\begin{tiny}
\begin{lstlisting}[breaklines=true,escapechar=$]
def func0(var0, var1, var2):
...
    var5 = var2
    var6 = []
    var7 = 0
\end{lstlisting}
\end{tiny}

Sometimes models are trying so "hard" to be helpful that they insert parts of instructions, instructions or even whole pieces of code that they weren't asked to or even specifically told not to. For example, llama models have a tendency to add code asking for user input. 

Below is an example where a model was asked to generate a new list of strings.

\begin{tiny}
\begin{lstlisting}[breaklines=true,escapechar=$]
Define a function ...
...
Assign a new list of strings to var7.
Assign 0 to var6.
\end{lstlisting}
\end{tiny}

The result code was this:

\begin{tiny}
\begin{lstlisting}[breaklines=true,escapechar=$]
def __main__(var0):
    var7 = []
    $\colorbox{yellow}{var7 = ["Hello", "World"]}$
    var6 = 0
...
\end{lstlisting}
\end{tiny}

A final example, an error that was quite common a few months ago, but seems to have mostly been solved recently, is mismatched parenthesis.
This error is trivial to notice since the code won't parse/compile.
Our dataset has many redundant parenthesis because instructions are generated in a single pass and we want to ensure correct operator order.
Here is an example were many parenthesis caused the model to err:

\begin{tiny}
\begin{lstlisting}[breaklines=true,escapechar=$]
Define a function ...
...
While var8 is less than var2 do the following and increment var8:
	 1. Assign min(var9, (var7 minus 1)) to var9.
	 2. While ((var9 plus 1) is less than var7) and ((array_index(var5, array_index(var6, (var9 plus 1))) plus (array_index(var1, array_index(var6, (var9 plus 1))) multiplied by array_index(var0, var8))) is less than or equal to (array_index(var5, array_index(var6, var9)) plus (array_index(var1, array_index(var6, var9)) multiplied by array_index(var0, var8)))) do:
...
\end{lstlisting}
\end{tiny}

As can be seen, there are a lot of parenthesis and clearly some are redundant, for example around \textit{(var9 plus 1)}.

\begin{tiny}
\begin{lstlisting}[breaklines=true,escapechar=$]
def __main__(var0, var1):
...    
    while var8 < var2:
        var9 = min(var9, (var7 - 1))
        while (var9 + 1 < var7) and $\colorbox{yellow}{\\(}$(var5[var6[var9 + 1]] + (var1[var6[var9 + 1]] * var0[var8])) <= (var5[var6[var9]] + (var1[var6[var9]] * var0[var8])):
...
\end{lstlisting}
\end{tiny}

The marked parenthesis does not have a matching closing one.

\section{Appendix B: The Prompt}
\label{appendix:prompt}

The general instructions to the models started from the simplest possible instruction, where LANGUAGE was Python.

{\small
\texttt{
\newline
Implement the following pseudocode in {LANGUAGE}.
\newline
}}

However, that was not enough for any of the models. They would not implement most of the function calls most of the time, with the exception of GPT which frequently implemented them, or at least said they should be implemented. 
Therefore, this was added:

{\small
\texttt{
\newline
The implementation should cover all aspects of the provided pseudocode, leaving no functions or functionality unimplemented.
\newline
}}

This was enough for gpt, but all other models required more detailed instructions.
Since they all seem to understand \textit{*} as \textit{any text} we defined these:

{\small
\texttt{
\newline
Replace all array\_* methods with {LANGUAGE} list operations\newline
Replace all string\_* methods, substring* and concat with {LANGUAGE} string operations.\newline
}}

Some instructions are \textbf{\textit{very}} specific to a certain issue that was common:
Reminding the model to update loop variable reduced those errors by half.
Telling it to leave containers empty unless specifically requested reduced some of the hallucinations. 
The instruction to declare global variables outside the function helped granite models and llama-2, but hurt llama-3 badly. Llama-3 would randomly choose a declared variable and place it in global scope without properly declaring it. Removing that instruction almost completely solved the issue.

{\small
\texttt{
\newline
Update loop variable before issuing the "continue" keyword.\newline
Global variables must be outside the function.\newline
Unless specifically requested, initialization is to an empty container.\newline
}}

The end result is this set of instructions. These were part of the prompt with which we ran the tests. Additional model-specific template related text had to be added for llama and granite. 

{\small
\texttt{
\newline
Implement the following pseudocode in {LANGUAGE}.\newline
The implementation should cover all aspects of the provided pseudocode, leaving no functions or functionality unimplemented.\newline
Do NOT ask for user input.\newline
Always define the requested function!\newline
Replace all array\_* methods with {LANGUAGE} list operations\newline
Replace all string\_* methods, substring* and concat with {LANGUAGE} string operations.\newline
Update loop variable before issuing the "continue" keyword.\newline
Do NOT add any additional text. Wrap the code in triple back-quotes.\newline
Global variables must be outside the function.\newline
Unless specifically requested, initialization is to an empty container.\newline
}}

\section{Appendix C: Common Errors}
\label{appendix:error_labels}

The explanation of the errors is as follows:
{
\begin{itemize}[nosep]
    \item \textbf{loop}: Usually missing loop variable update, mostly resulting in infinite loops (or array OOB); sometimes misplaced loop variable update
    \item \textbf{ignored}: Some instruction, or part of, was not implemented
    \item \textbf{wrong}: The implementation is different than required, e.g. reversed assignment, used wrong variable. Could also call this one hallucinations.
    \item \textbf{ASCII}: Errors interpreting strings as ints as done in C/C++, mostly missing ord() or chr() in some of the program instructions
    \item \textbf{unbalanced}: The most common parsing/compilation error, unbalanced parenthesis
    \item \textbf{division}: Several tests require integer results from division requiring a special operator
    \item \textbf{indent}: Using the wrong indentation for the line of code
    \item \textbf{split}: Many tests require splitting a string (by whitespace) into an array often misinterpreted by the model. Often split is actually redundant.
    \item \textbf{global}: If modifying a global variable it needs to be declared; this is sometimes missing
    \item \textbf{other}: Any error that doesn't fit the common categories, e.g. running out of tokens
\end{itemize}
}

\end{document}